\documentclass[runningheads,a4paper]{llncs}

\usepackage[a4paper, margin=1.65in, top=1.5in,bottom=1.5in]{geometry}
\usepackage{amssymb}
\usepackage{graphicx}
\usepackage{multirow}
\usepackage{array}
\usepackage{color}
\usepackage{float}
\usepackage{enumerate}
\usepackage{amsmath}
\usepackage{multirow}
\usepackage{rotating}

\usepackage{url}

\long\def\comment#1{}

\long\def\comment#1{}

\usepackage{xspace}
\newcommand*{\eg}{e.g.\@\xspace}
\newcommand*{\ie}{i.e.\@\xspace}
\newcommand*{\etal}{et al.\@\xspace}

\mathchardef\mhyphen="2D

\usepackage{bbding,array}

\newcolumntype{M}[1]{>{\centering\arraybackslash}m{#1}}

\usepackage[sort&compress,square,comma,numbers]{natbib}
\makeatletter
\renewcommand\bibsection%
{
  \section*{\refname
    \@mkboth{\MakeUppercase{\refname}}{\MakeUppercase{\refname}}}
}
\makeatother
\usepackage{caption}
\usepackage{subcaption}

\begin{document}

\mainmatter  

\title{Semantic-guided Encoder Feature Learning for Blurry Boundary Delineation}

\titlerunning{Semantic-guided Encoder Feature Learning for Blurry Boundary Delineation}

%
%
\author{Dong Nie$^{1,2}$, Dinggang Shen$^{1,}$\thanks{Corresponding author.}%
}
\institute{
$^1$Department of Radiology and BRIC, University of North Carolina at Chapel Hill, USA\\
$^2$Department of Computer Science, University of North Carolina at Chapel Hill, USA\\
}
%

%
%

\tocauthor{Dong Nie, Dinggang Shen}
\maketitle

\begin{abstract}
Encoder-decoder architectures are widely adopted for medical image segmentation tasks. With the lateral skip connection, the models can obtain and fuse both semantic and resolution information in deep layers to achieve more accurate segmentation performance. However, in many applications (\eg, images with blurry boundary), these models often cannot precisely locate complex boundaries and segment tiny isolated parts due to the fuzzy information in the skip connection provided from the encoder layers. To solve this challenging problem, we first analyze why simple skip connections are not sufficient to help accurately locate indistinct boundaries. Then we propose a semantic-guided encoder feature learning strategy to learn high resolution semantic encoder features so that we can more accurately locate the blurry boundaries, which can also enhance the network by selectively learning discriminative features. Besides, we further propose a soft contour constraint mechanism to model the blurry boundary detection. Experimental results on real clinical datasets show that our proposed method can achieve state-of-the-art segmentation accuracy, especially for the blurry regions. Further analysis also indicates that our proposed network components indeed contribute to the performance gain. Experiments on extra dataset validate the generalization ability of our proposed method.
\end{abstract}

\section{Introduction}
\label{sec:intro}
\begin{figure*}[!htb]
\centering
  \includegraphics[width=1\linewidth]{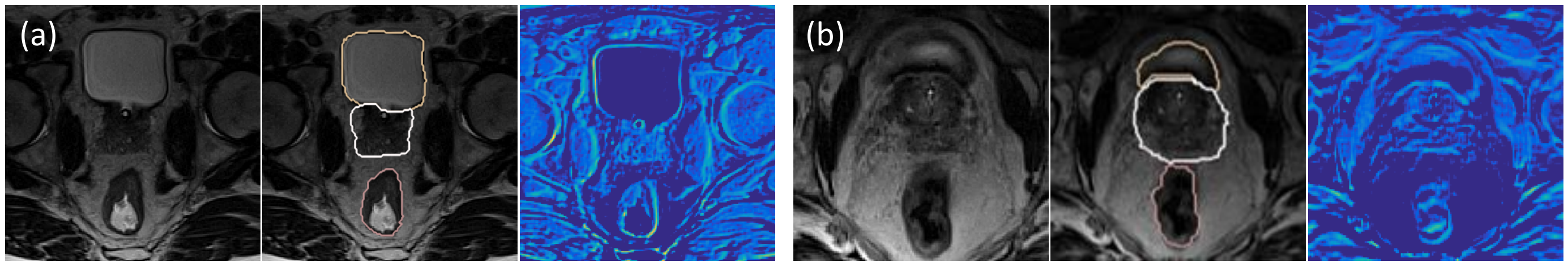}\\
  \caption{Illustration of the blurry and vanishing boundaries
within pelvic MRI images, together with overlaid ground truth contour and the typical feature maps in the encoder layer of a conventional UNet. (a) and (b) are the two typical slices of two subjects, in which boundaries of bladder and rectum are relatively clear, but prostate is blurry.}
\label{fig:example}
\vspace{-5pt}
\end{figure*}

\vspace{-5pt}
Automatic image segmentation is an essential step for many medical image analysis applications, include computer-aided radiation therapy, disease diagnosis and treatment effect evaluation. One of the major challenges for this task is the blurry nature of medical images (e.g., CT, MR and, microscopic images), which can often result in low-contrast and even vanishing boundaries, as shown in Fig.~\ref{fig:example}.

Many encoder-decoder based networks have been proposed for semantic segmentation~\cite{long2015fully,ronneberger2015u,yu2017volumetric} and achieved very promising performances on various tasks. UNet~\cite{ronneberger2015u}, a typical encoder-decoder architecture which combines shallow and deep features with a skip connection is widely used in many image segmentation tasks. Some works are proposed to enhance the UNet~\cite{roy2018concurrent,oktay2018attention,zhou2018fine}. However, Heller~\etal~\cite{heller2018imperfect} found that the deep segmentation models are robust on the non-boundary regions, but not very robust to boundaries.
Actually, these models usually fail to properly segment the blurry boundaries, especially for the case with extremely low tissue contrast. For example, prostate boundaries in MR or CT pelvic images are often blurry. To solve this challenge, we argue high resolution with rich semantic based feature learning is desired.

Besides the variants of UNet, to better delineate the boundaries, Ravishankar~\etal~\cite{ravishankar2017joint} proposed a multi-task network to robustly segment the organs by jointly regressing the boundaries and foreground. Zhu~\etal~\cite{zhu2019boundary} proposed a boundary-weighted domain adaptive neural network to accurately extract the boundaries of MRI prostate. However, all these methods do not consider the fact that the voxels around blurry boundaries are highly similar. Thus, we should not directly classify or regress the voxels to be on the boundary or not.

In this paper, we propose a novel semantic-guided encoder feature learning mechanism to improve the skip connection in previous encoder-decoder architectures, so that we can work better for low-contrast medical image segmentation. The design of our network is mainly based on the idea of explicitly utilizing high-resolution semantic information to compensate for the deficiency on inaccurate boundary delineation of the existing encoder-decoder networks. Specifically, we propose to concatenate the low-layer (encoder) feature maps and the high-layer (decoder) feature maps, and then design a channel-wise attention and spatial-wise attention to help learn (which can also be viewed as a kind of feature selection) the high-resolution semantic encoder feature maps. With these better learned encoder feature maps, we further concatenate (or element-wisely add) it to the corresponding decoder layers in encoder-decoder framework. Moreover, we propose using soft label to indicate the probability of a voxel being on the boundary. Accordingly, a soft cross-entropy loss is proposed as a metric for the blurry boundary delineation problem.

\vspace{-5pt}
\section{Method}
\label{sec:method}
\begin{figure*}[!htb]
\centering
  \includegraphics[width=0.6\linewidth]{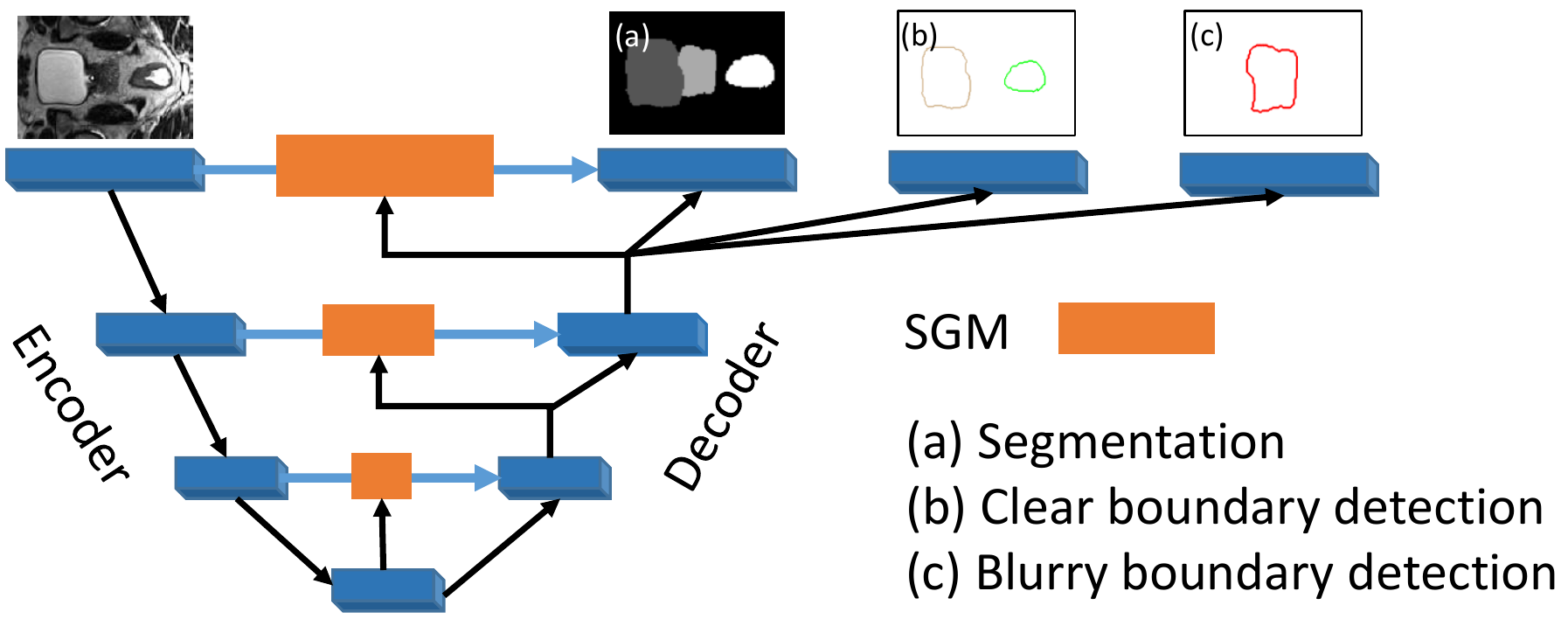}\\
  \caption{Illustration of the architecture of our proposed method, which consists of a semantic-guided module (SGM). (a) means a segmentation branch, and (b) and (c) indicate boundary detection branches.}
\label{fig:asd_arch}
\vspace{-10pt}
\end{figure*}

\vspace{-5pt}
The architecture of our proposed framework is presented in Fig.~\ref{fig:asd_arch}, in which an encoder-decoder architecture is introduced with three tasks (segmentation, clear boundary detection, and blurry boundary detection). The proposed semantic-guided encoder feature learning module (SGM) is further highlighted in Fig.~\ref{fig:semantic_encoding}. 


In the following subsections, we will analyze the deficiency of the skip connection in the current encoder-decoder framework. Then, we introduce the proposed semantic-guided encoder feature learning strategy. Moreover, we will describe the soft contour constraint for blurry boundary delineation. Finally, we describe the implementation details.

\vspace{-5pt}

\subsection{Analysis of Skip Connection in Encoder-Decoder Architecture}
In the classical encoder-decoder architecture~\cite{ronneberger2015u}, shallow and deep features are usually complementary to each other. For example, shallow features are rich in resolution but insufficient in semantic information, while deep features are highly semantic but lack of spatial details. The skip connection proposed in UNet~\cite{ronneberger2015u} is supposed to provide high-resolution information from the shallow (encoder) layers to the deep (decoder) layers, so that we can improve localization precision without losing classification accuracy. However, the raw (simple) skip connection has several drawbacks. a) It would bring `noise' (unnecessary information) to the deep layers which will definitely affect the concatenation of feature maps, as shown in the visualized encoder feature maps in Fig.~\ref{fig:example}. b) The huge gap between shallow and deep features will decrease
the power of this combination. c) Moreover, for the clear boundaries (\eg, bladder and rectum), the encoder feature maps can provide sufficiently precise localization information as shown in Fig.~\ref{fig:example}, which can thus work well with the raw skip connection. But the blurry boundary (\eg, prostate) cannot be well described in the encoder feature maps as shown in Fig.~\ref{fig:example}, which thus cannot provide accurate localization information with simple skip connection. Therefore, it is highly desired to select discriminative features, not simply inhibiting indiscriminative features from shallow layers; in other words, we should learn high resolution semantic features from the encoder. To achieve such an effect, Roy \etal~\cite{roy2018concurrent} proposed concurrent spatial-and-channel squeeze and excitation module to boost meaningful features and suppress weak ones. Oktay~\etal~\cite{oktay2018attention} proposed gated attention mechanism to select the salient part of the feature maps to further improve the UNet. However, in both works, the feature learning process is actually conducted in an \textit{implicit} manner which will limit the learning efficiency.
\vspace{-10pt}

\subsection{Semantic-guided Encoder Feature Learning}
\label{subsec:sg}
\begin{figure*}[!htb]
\centering
  \includegraphics[width=0.75\linewidth]{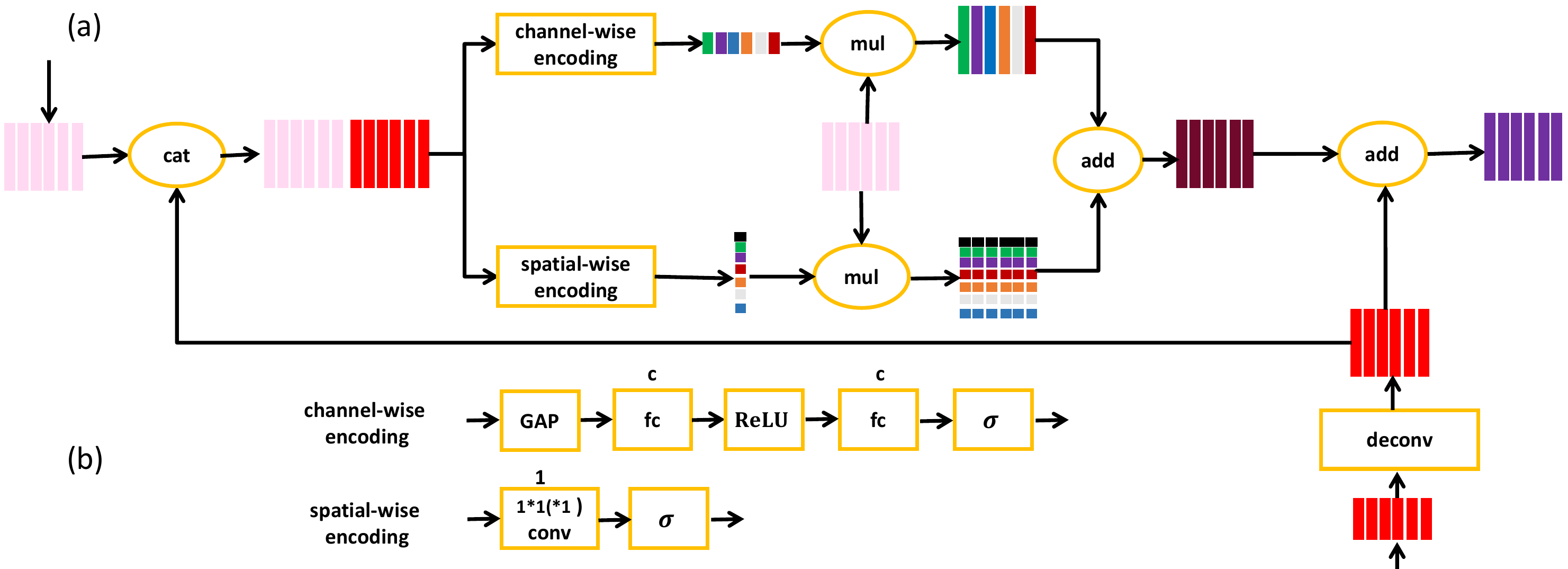}\\
  \caption{Illustration of the proposed semantic-guided module (SGM), as shown in ( a ). The pink blocks represent the features of shallow layers, while the red ones represent the features of deep layers. Different from direct skip connection in UNet, we propose using semantic concept from deep layers to guide feature learning in the corresponding shallow layers, for which a channel-wise encoder and a spatial-wise encoder are both proposed, as shown in (b). `GAP' means Global Average Pooling.}
\label{fig:semantic_encoding}
\end{figure*}

\vspace{-5pt}
To overcome the above mentioned problems, we propose to \textit{explicitly} learn the high resolution semantic features (which are also more discriminative) from shallow (encoder) layers with semantic guidance from deep (decoder) layers. The key idea is to encode semantic concept from deep layer features to guide the learning of shallow features.
As shown in Fig.~\ref{fig:semantic_encoding}, our semantic-guided feature learning module (\ie, SG module or SGM) is designed to selectively enhance or suppress the features of shallow layer at each stage so that we can enhance the consistency between shallow and deep layers without losing resolution information. Besides the widely-used channel-wise encoder, we have also designed the spatial-wise encoder as described below.

We consider the feature maps of a certain encoder layer (\ie, shallow features) to be \(S = \left\{ {{s_1},{s_2},...,{s_K}} \right\}\), where \({s_i} \in {R^{H \times W \times T}}\). We also assume the up-sampled feature maps in the corresponding decoder layer (deep features) to be \(D = \left\{ {{d_1},{d_2},...,{d_K}} \right\}\), where \({d_i} \in {R^{H \times W \times T}}\). We concatenate the two group of feature maps together and thus result in a bank of \textit{high-resolution and rich-semantic mixed} feature maps as shown in Eq.~\ref{eq:cat}.
\begin{equation}
F = \left\{ {{s_1},{s_2},...,{s_K},{d_1},{d_2},...,{d_K}} \right\}.
\label{eq:cat}
\end{equation}

\noindent{\bf Channel-wise Encoding:}
With a global average pooling layer, we obtain a vector \(Q = \left\{ {{q_1},{q_2},...,{q_K},...{q_{2K}}} \right\}\), where $q_k$ is a scalar and corresponds to the averaging value of the k-th feature maps in $F$. Then, two successive fully connected layer are adopted to fuse the resolution and semantic information: \(Z = {W_1}\left( {\mathop{\rm ReLU} \left( {{W_2}Q} \right)} \right)\), with \({W_1} \in { R^{K \times K}}\) and \({W_2} \in { R^{2K \times K}}\).
This encodes the channel-wise dependencies by considering both shallow and deep features. We apply a sigmoid activation function to map the neurons to probabilities so that we can formulate as a channel-wise importance descriptor, which can be described as \(\sigma \left( Z \right)\).
Thus, the semantic-guided channel-wise encoded feature maps are formulated as Eq.~\ref{eq:sgcf}.
\begin{equation}
SGCF = \left\{ {\sigma \left( {{z_1}} \right){s_1},\sigma \left( {{z_2}} \right){s_2},...,\sigma \left( {{z_K}} \right){s_K}} \right\}
\label{eq:sgcf}
\end{equation}

Note that the weight \(\sigma \left( {{z_k}} \right)\) before the shallow feature map \(s_k\) can be viewed as an indicator of how important this specific feature map is. Thus, we argue this channel-wise encoding is actually a semantic-guided feature selection process in a channel-wise manner, which is able to ignore less meaningful feature maps and emphasize the meaningful ones. In other words, it can help remove the `noise' and retain the useful information. More importantly, since \(\sigma \left( Z \right)\) has taken both high resolution and rich semantic information into account, it has more discriminative capacity than the case of only considering shallow layer information in~\cite{roy2018concurrent}.

\noindent{\bf Spatial-wise Encoding:}
Now we come to consider the spatial-wise importance to achieve better fine-grained image segmentation.

Based on the concatenated feature maps $F$, we apply a \(2K \times 1 \times 1 \times 1\) convolution to squeeze the channels. Therefore, we can obtain a one-channel output feature map $U$, where \(U \in {R^{H \times W \times T}}\). We directly apply sigmoid function to acquire a probability map for $U$. Similarly, the semantic-guided spatial-wise encoded shallow feature maps can be described in Eq.~\ref{eq:cesfm}.
\begin{equation}
SGSF = \left\{ {\sigma \left( U \right) \otimes {s_1},\sigma \left( U \right) \otimes {s_2},...,\sigma \left( U \right) \otimes {s_K}} \right\}
\label{eq:cesfm}
\end{equation}
Since \(\sigma \left({U_{h,w,t}}\right)\) corresponds to the relative importance of a spatial information at \(\left( {h,w,t} \right)\) of a given shallow layer feature map, it can help select more important features to relevant spatial locations and also ignore the irrelevant ones. Moreover, \(\sigma \left( U \right)\) is a fusion of both resolution and rich semantic information, thus it can provide a better localization capacity even for the blurry boundary regions which cannot done by~\cite{roy2018concurrent}. As a result, we view this spatial-wise encoding as a semantic-guided recalibration process.

\noindent{\bf Combination of Encoded Feature Maps:}
Now we can formulate both channel-wise and spatial-wise encoding by a simple element-wise addition operation, as shown in eq.~\ref{eq:esfm}.
\begin{equation}
SGF = {\rm{SGCF}} + {\rm{SGSF}}
\label{eq:esfm}
\end{equation}
This \(SGF\) considers both channel-wise encoded and spatial-wise encoded information, thus, it contains \textit{not only} the discriminative (semantic) features, \textit{but also} more accurate localization information.




\noindent{\bf Final Combination with Deep-Layer Feature Maps:}
To this end, we can simply complete the concatenation operation or element-wise addition operation. Instead of using the shallow feature maps $S$, we use the channel-wise and spatial-wise encoded shallow feature maps $SGF$ to combine with the deep-layer feature maps $D$ (through concatenation or element-wise addition). 
Compared with the raw skip connection in UNet, our encoded shallow feature maps $SGF$ has same resolution but much more semantic and precise localization information (especially for the blurry regions), and thus can make the combination more reasonable. At the same time, since the operations in the encoder are mostly \(1 \times 1 \times 1\) convolution, the number of parameters just increases a little bit.

To further increase the model's discriminative capacity, we also adopt the multi-scale deep supervision strategy as in~\cite{yu2017volumetric} after feature fusion at each stage.
\vspace{-5pt}

\subsection{Boundary Delineation with Soft Contour Constraint}
\label{subsec:contour}
\vspace{-5pt}
In mammal visual system~\cite{lamme1999separate}, contour delineation closely correlates with object segmentation.
To incorporate the knowledge to improve the segmentation accuracy, we integrate the task of contour detection with the task of segmentation, assuming that introducing a task of contour detection can help guide the network to concentrate more on the boundaries of organ regions, thus helping overcome the adverse effect of low tissue contrast.
In this paper, as shown in Fig.~\ref{fig:asd_arch}, two boundary detection tasks are added to the end of the network as auxiliary guidance.

To extract the contour for training, we first delineate the boundaries of different organs by performing Canny detector 
on the ground-truth segmentation. For the organs with clear boundaries (\ie, bladder and rectum in our case), we model the problem as a classification problem. However, due to the potential sample imbalance problem, we propose using focal loss to alleviate such an issue, as shown in Eq.~\ref{eq:cboundary}.
\begin{equation}
{L_{cboundary}} =  - \sum\nolimits_h {\sum\nolimits_w {\sum\nolimits_t {\sum\nolimits_{c \in csets} {{I_{\left\{ {{Y_{h,w,t}},c} \right\}}}{{\left( {1 - \hat p\left( {{X_{h,w,t}};\theta } \right)} \right)}^\gamma }\left( {1 - \hat p\left( {{X_{h,w,t}};\theta } \right)} \right)} } } }
\label{eq:cboundary}
\end{equation}

Note that, for the regions with blurry boundaries (\ie, prostate in our case), the voxels near the boundaries look almost same. As a result, it will be more reasonable to assign soft labels (instead of hard labels) around the ground-truth boundaries. Thus, we can formulate the blurry-boundary delineation task as a soft classification problem, which estimates the probability of each voxel being on the organ boundaries. Then, for these blurry boundaries, we further exert a Gaussian filter (with a bandwidth of \(\delta \), \ie, empirically set to $3$ in our study) on the obtained boundary map. In other words, for each voxel, we generate an approximate probability belonging the blur boundary of an organ. Hence, we can formulate soft classification as a soft cross-entropy loss function as defined in Eq.~\ref{eq:bboundary}.
\begin{equation}
{L_{bboundary}} =  - \sum\nolimits_h {\sum\nolimits_w {\sum\nolimits_t {{p_{h,w,t}}\left( {1 - \hat p\left( {{X_{h,w,t}};\theta } \right)} \right)} } }
\label{eq:bboundary}
\end{equation}
\vspace{-15pt}
\subsection{Implementation Details}
\label{subsec:imp}
\vspace{-3pt}
Pytorch\footnote{https://github.com/pytorch/pytorch} is adopted to implement our proposed method shown in Fig.~\ref{fig:asd_arch}.
The code can be obtained by this link\footnote{https://github.com/ginobilinie/SemGuidedSeg}.
We adopt Adam algorithm to optimize the network. The input size of the segmentation network is \(144 \times 144 \times 16\). The network weights are initialized by the Xavier algorithm,
and weight decay is set to be 1e-4. For the network biases, we initialize them to 0. The learning rate for the network is initialized to 2e-3, followed by decreasing the learning rate 10 times every 2 epochs during the training until 1e-7. Four Titan X GPUs are utilized to train the networks.

\section{Experiments and Results}
\label{sec:results}
\vspace{-5pt}
Our pelvic dataset consists of 50 prostate cancer patients from a cancer hospital, each with one T2-weighted MR image and corresponding manually-annotated label map by medical experts. In particular, the prostate, bladder and rectum in all these MRI scans have been manually segmented, which serve as the ground truth for evaluating our segmentation method. All these images were acquired with 3T MRI scanners.
The image size is mostly $256\times256\times\left( {120 \sim 176} \right)$, and the voxel size is $1\times1\times1~\text{mm}^3$.
A typical example of the MR image and its corresponding label map are given in Fig.~\ref{fig:example}.

Five-fold cross validation is used to evaluate our method. Specifically, in each fold of cross validation, we randomly chose 35 subjects as training set, 5 subjects as validation set, and the remaining 10 subjects as testing set.
Unless explicitly mentioned, all the reported performance by default is evaluated on the testing set. As for evaluation metrics, we utilize Dice Similarity Coefficient (DSC) and Average Surface Distance (ASD) to measure the agreement between the manually and automatically segmented label maps.

\vspace{-8pt}
\subsection{Comparison with State-of-the-art Methods}
\label{subsec:comarison}
\vspace{-5pt}
To demonstrate the advantage of our proposed method, we also compare our method with other three widely-used methods on the same dataset as shown in Table~\ref{tab:res}: 
1) SSAE~\cite{guo2016deformable},
2) UNet~\cite{ronneberger2015u}, 
3) SResSegNet~\cite{yu2017volumetric}. 
\begin{figure}[!htb]
\vspace{-20pt}
\centering
  \includegraphics[width=0.75\linewidth]{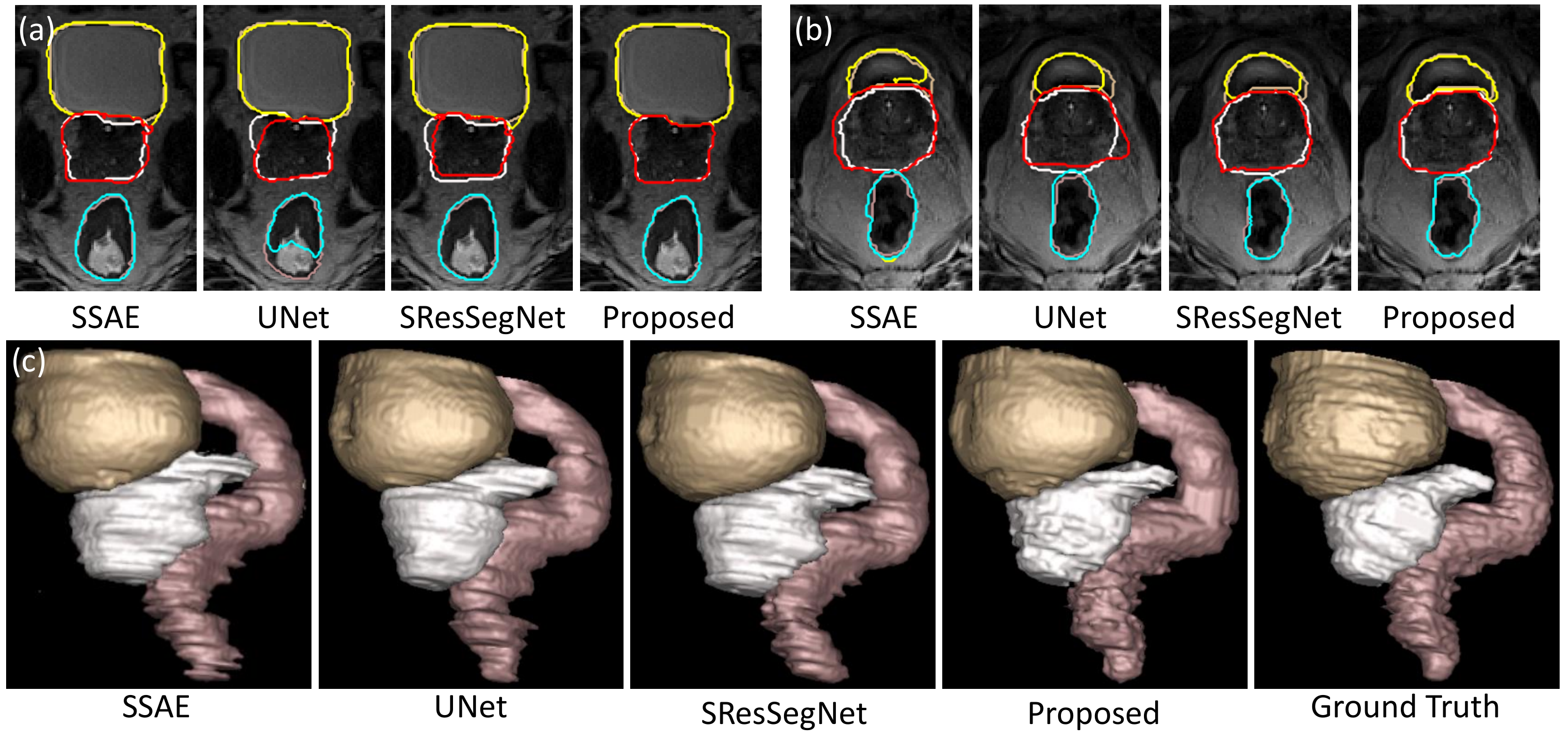}\\
  \caption{Visualization of pelvic organ segmentation results by four methods. In (a) and (b), orange, silver and pink contours indicate the manual ground-truth segmentations, while yellow, red and cyan ones indicate automatic segmentations. (a) Clear boundary case, (b) blurry boundary case, and (c) 3D renderings of segmentations.}
\label{fig:res_comparison}
\vspace{-10pt}
\end{figure}

\begin{table*}[!htbp]
\centering
\vspace{-30pt}
\caption{DSC and ASD on the pelvic dataset by four different methods.}
\begin{tabular}{l ccccccc}
\hline
\multirow{2}{*}{Method} & \multicolumn{3}{c}{DSC} & \multicolumn{3}{c}{ASD}\\
                         & Bladder & Prostate & Rectum & Bladder & Prostate & Rectum\\
\hline
SSAE & .918(.031)  & .871(.042) &  .863(.044)&   1.089(.231)  & 1.660(.490) & 1.701(.412)\\
UNet &.896(.028) & .822(.059) &  .810(.053)&  1.214(.216) & 1.917(.645) &2.186(0.850)\\
SResSegNet &.944(.009) & .882(.020) & .869(.032)&  .914(.168) & 1.586(.358) &1.586(.405)\\
Proposed & {\bf .975(.006)} & {\bf .932(.017)}& \textbf{.918(.025)} & {\bf .850(.148)}  &  {\bf 1.282(.273)} & \textbf{1.351(.347)}\\
\hline
\end{tabular}
\label{tab:res}
\vspace{-15pt}
\end{table*}
Table~\ref{tab:res} quantitatively compares our method with three state-of-the-art segmentation methods.
We can see that our method achieves better accuracy than the other state-of-the-art methods in terms of both DSC and ASD. It is worth noting that our proposed method can achieve much better performance for the blurry-boundary organ (\ie, prostate), which indicates the effectiveness of our proposed network components for blurry boundary delineation.

We also visualize some typical segmentation results in Fig.~\ref{fig:res_comparison}, which further show the superiority of our proposed method, especially for the blurry regions around the prostate.
\vspace{-8pt}

\subsection{Impact of Each Proposed Component}
\label{subsec:analysis}
\vspace{-5pt}
As our method consists of several novel proposed components, we conduct empirical studies below to analyze them.

\noindent\textbf{Impact of Proposed SG Module:}
As mentioned in Sec.~\ref{subsec:sg},
we propose a semantic-guided encoder feature learning module to learn more discriminative features in shallow layers. The effectiveness of the SG module is further confirmed by the improved performance, \eg, 2.40\%, 4.41\% and 2.8\% performance improvements in terms of DSC for bladder, prostate, and rectum, respectively, compared with the UNet with multi-scale deep supervision.

\begin{figure}[!htb]
\vspace{-15pt}
\centering
  \includegraphics[width=1\linewidth]{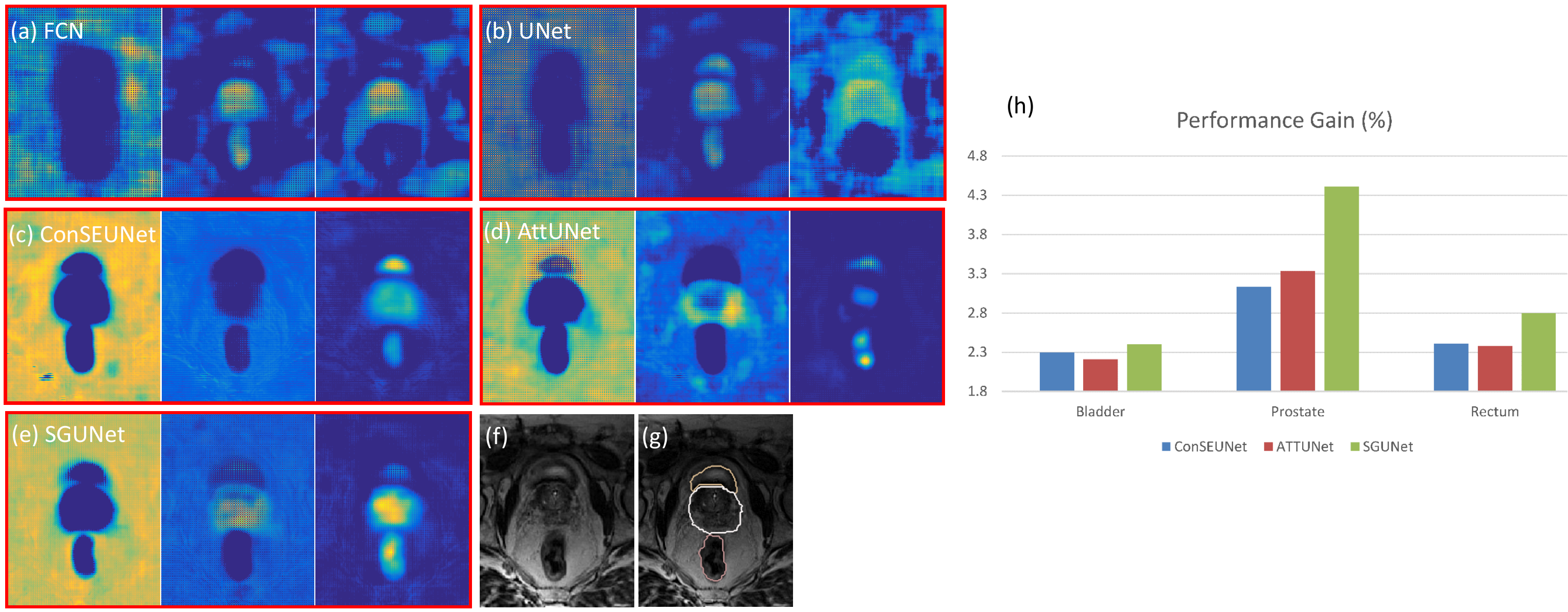}\\
  \caption{(a-e): Visualization of three typical learned feature maps of a certain layer by five different networks. (f) and (g) are the corresponding input MRI and the MRI overlaid by the ground-truth contours. (h) is the performance gain with different strategies towards the UNet with \textit{multi-scale deep supervision}.}
\label{fig:feature_maps}
\vspace{-15pt}
\end{figure}

\noindent\textbf{Relationship with Similar Work:} Several previous work are proposed to use attention mechanism~\cite{roy2018concurrent,oktay2018attention} to enhance the encoder-decoder networks. However, our work is different from them mainly in the follow way: We propose to use highly semantic information from the decoder to explicitly guide the building of attention mechanism, so that we can efficiently learn the encoder features. To further compare them, we visually present the three \textit{typical} learned feature maps (selected by clustering) of a certain layer (\ie, combined layer) in different networks at a certain training iteration (\ie, 4 epochs). The methods include FCN~\cite{long2015fully}, UNet~\cite{ronneberger2015u}, UNet with concurrence SE module~\cite{roy2018concurrent} (ConSEUNet), attention-UNet~\cite{oktay2018attention} (AttUNet) and our proposed one(SGUNet). The visualized maps are in Fig.~\ref{fig:feature_maps}.

\vspace{-5pt}

Fig.~\ref{fig:feature_maps}(a-e) indicates that the raw encoder-decoder network (\ie, FCN and UNet) cannot handle well for the blurry boundary cases. The attention based networks can generate higher semantic maps with better localization information. Among them, our proposed method can learn more precise boundaries due to explicit semantic guidance. Also, our proposed method have a faster convergence compared to other methods. Besides, the quantitative analysis in Fig.~\ref{fig:feature_maps}(h) is consistent with the the conclusion of qualitative analysis.

\noindent\textbf{Impact of Soft Contour Constraint:}
As introduced in Sec.~\ref{subsec:contour}, we apply a hard contour constraint for clear-boundary organs while a soft contour constraint for the blurry-boundary organs. Since hard contour constraint is a widely adopted strategy, we directly compare our proposed soft contour constraint with the case of using hard constraint. With soft constraint on the prostate, we can achieve a slight performance gain such as 0.2\% in terms of DSC; but we can achieve more performance gain in terms of ASD (0.8\%), which is mainly because the soft contour constraint can help more accurately locate the blurry boundaries.
\vspace{-8pt}
\subsection{Validation on Extra Dataset}
\label{lab:challenge}
\vspace{-3pt}
To show the generalization ability of our proposed algorithm, we conduct additional experiments on the PROMISE12-challenge dataset~\cite{litjens2014evaluation}. This dataset contains 50 labeled subjects where only prostate was annotated. 
We can achieve a high DSC ($0.92$), small ASD ($1.57$) in average based on five-fold cross validation. As for the extra 30 subjects' testing dataset whose ground-truth label maps are hidden from us, the performance of our proposed algorithm is still very competitive (we are ranking in the top 6 among 290 submission with an overall score of $89.46$. The details can be available via this link\footnote{https://promise12.grand-challenge.org/evaluation/results/}) compared to the state-of-the-art methods on the 30 subjects' testing dataset~\cite{yu2017volumetric,zhu2019boundary}. These experimental results indicate a very good generalization capability of our proposed algorithm.
\vspace{-5pt}
\section{Conclusion}
\label{sec:conclusion}
\vspace{-5pt}
In this paper, we have presented a novel semantic-guided encoder feature learning strategy to learn both highly semantic and rich resolution information features, so that we can better deal with the blurry-boundary delineation problem. In particular, our SG module can improve the raw skip connection of the raw encoder-decoder models by enhancing the discriminative features while compressing the less informative features. Furthermore, we propose a soft contour constraint to model the blurry-boundary detection, while an ordinary hard contour constraint to model the clear-boundary detection; this strategy is validated effective to help boundary localization and alleviate inter-class errors. By integrating all these proposed components into the network, our final proposed framework has achieved sufficient improvement compared to other methods, in terms of both accuracy and robustness, also on the extra dataset.
\vspace{-10pt}
\comment{
\section*{Acknowledgments}
The authors gratefully acknowledge the generous support from National High-tech R\&D Program of China (2013AA01A606),
NSFC (61070115), Institute of Psychology (113000C037),
Strategic Priority Research Program (XDA06030800) and 100-Talent Project (Y2CX093006) from Chinese Academy of Sciences.
}

\end{document}